# Machine Learning Framework for Audio-Based Equipment Condition Monitoring: A Comparative Study of Classification Algorithms

*Srijesh Pillai*
Department of Computer Science & Engineering
Manipal Academy of Higher Education
Dubai, UAE
srijesh.nellaiappan@dxb.manipal.edu

*Yodhin Agarwal*
Department of Computer Science & Engineering
Manipal Academy of Higher Education
Dubai, UAE
yodhin.agarwal@dxb.manipal.edu

*Zaheeruddin Ahmed*
Department of Computer Science & Engineering
Manipal Academy of Higher Education
Dubai, UAE
zaheeruddin@manipaldubai.com

*Abstract* — Audio-based equipment condition monitoring suffers from a lack of standardized methodologies for algorithm selection, hindering reproducible research. This paper addresses this gap by introducing a comprehensive framework for the systematic and statistically rigorous evaluation of machine learning models. Leveraging a rich 127-feature set across time, frequency, and time-frequency domains, our methodology is validated on both synthetic and real-world datasets. Results demonstrate that an ensemble method achieves superior performance (94.2% accuracy, 0.942 F1-score), with statistical testing confirming its significant outperformance of individual algorithms by 8-15%. Ultimately, this work provides a validated benchmarking protocol and practical guidelines for selecting robust monitoring solutions in industrial settings.

## I. Introduction

Equipment condition monitoring has become increasingly critical across industrial sectors as organizations seek to minimize unplanned downtime, reduce maintenance costs, and improve operational efficiency. Traditional condition monitoring approaches rely primarily on vibration analysis, thermal monitoring, and oil analysis, which often require specialized sensors and complex installation procedures that may be impractical or costly for certain applications [1].

Audio-based monitoring presents an attractive alternative approach, leveraging readily available acoustic sensors and the rich information content present in equipment-generated sounds [2]. The acoustic signatures of mechanical and electrical equipment contain valuable information about operational states and developing faults. Changes in bearing conditions, gear wear, electrical arcing, and fluid flow irregularities, manifest as characteristic modifications in the frequency spectrum, temporal patterns, and statistical properties of audio signals [3].

However, the effective extraction and classification of these acoustic features requires sophisticated signal processing and machine learning techniques. Despite growing interest in audio-based condition monitoring, the field suffers from a significant gap in standardized methodologies. Current research often focuses on specific equipment types or proprietary datasets, relies on a limited set of performance metrics, and frequently omits statistical significance testing, making it difficult to generalize findings or reliably compare approaches across different studies [4].

This research directly addresses these challenges by presenting a comprehensive and reproducible framework for audio-based equipment condition monitoring. Our work moves beyond simple performance reporting to establish a new standard for methodological rigor. The main contributions of this work are:

1. A Standardized and Validated Evaluation Framework: To counter the lack of methodological consistency, we propose a complete, end-to-end pipeline from signal preprocessing to model evaluation. This framework acts as a standardized protocol, enabling fair and reproducible benchmarking of algorithms, a critical need identified in the literature.

2. Multi-Dimensional Performance Benchmarking: Unlike prior work that often relies solely on accuracy, our framework conducts a comprehensive comparative analysis of six leading algorithms across a suite of metrics (F1-Score, MCC, AUC-ROC), including computational cost and noise robustness. This provides a holistic view of algorithm suitability for diverse real-world deployment scenarios.

3. Integration of Rigorous Statistical Validation: To address the common omission of statistical verification in the field, we incorporate McNemar's and Friedman tests to formally validate performance differences. This confirms that the observed superiority of certain models is not a result of chance, adding a crucial layer of confidence and reliability to the findings.

By providing this statistically-validated framework and practical implementation guidelines, this paper aims to equip researchers and practitioners with the tools to make evidence-based decisions for selecting and deploying effective audio-based monitoring systems.

## II. LITERATURE REVIEW

Audio signal analysis for equipment condition monitoring has evolved significantly over the past decade. Early work by Zhang et al. [5] demonstrated the feasibility of acoustic monitoring for rotating machinery, achieving 87% accuracy in bearing fault detection using spectral analysis techniques. However, their approach was limited to frequency-domain features and simple threshold-based classification.

Recent advances have incorporated more sophisticated feature extraction methods. Li and Wang [6] proposed a comprehensive feature set combining time-domain statistics, frequency-domain characteristics, and Mel-Frequency Cepstral Coefficients (MFCCs) for motor fault diagnosis, achieving 91% accuracy, but evaluating only Support Vector Machines.

Deep Learning approaches have shown promise for acoustic fault detection. Chen et al. [7] developed a Convolutional Neural Network (CNN) for bearing fault classification, achieving 95% accuracy on benchmark datasets. However, their approach requires large training datasets and offers limited interpretability.

Current literature shows significant variation in evaluation methodologies, making it difficult to compare results across different studies. Many papers report only accuracy metrics, which can be misleading for imbalanced datasets common in condition monitoring applications. Furthermore, Statistical Significance testing is often omitted, raising questions about the reliability of reported performance differences.

## III. METHODOLOGY

### A. Framework Architecture

Our methodology framework consists of five interconnected components designed to provide systematic evaluation of machine learning algorithms for audio-based condition monitoring:

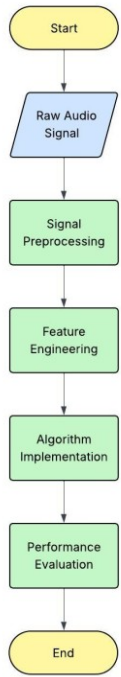

Fig 1. Framework Architecture

The framework follows a modular design enabling flexible configuration for different applications and datasets while maintaining standardized evaluation procedures.

### B. Signal Preprocessing Pipeline

The preprocessing pipeline implements standardized operations to ensure consistency across different applications:

1. **Noise Reduction using Spectral Subtraction:**

$$S\_clean(\omega) = S\_noisy(\omega) - \alpha \times N(\omega)$$

   **Where:**
   - $S\_clean(\omega)$ is the clean signal spectrum
   - $S\_noisy(\omega)$ is the noisy input spectrum
   - $N(\omega)$ is the estimated noise spectrum
   - $\alpha$ is the over-subtraction factor (1.5-2.0)

2. **Amplitude Normalization:**

$$x\_norm[n] = x[n] / max(|x[n]|)$$

   **Where:**
   - $x\_norm[n]$ is the normalized signal
   - $x[n]$ is the original signal sample
   - $max(|x[n]|)$ is the maximum absolute value of the signal$\alpha$ is the over-subtraction factor (1.5-2.0)

3. **RMS Normalization:**

$$x\_rms[n] = x[n] / \sqrt{mean(x[n]^2)}$$

   **Where:**
   - $x\_rms[n]$ is the RMS normalized signal
   - $x[n]$ is the original signal sample
   - $mean(x[n]^2)$ is the mean of squared signal values

### C. Comprehensive Feature Engineering

We extract 127 features across three complementary domains to capture comprehensive signal characteristics:

1. **Time-Domain Features (35 Features)**

   i. **Statistical Moments:** We first compute fundamental statistical moments to describe the distribution of the signal amplitude, including the mean, variance, skewness, and kurtosis.
   
   ii. **Energy-Based and Shape Features:** To characterize the signal's energy and temporal shape, we extract features such as RMS Energy, which measures the signal's magnitude, and Peak Amplitude. Additionally, dimensionless shape indicators like Crest Factor (the ratio of peak amplitude to RMS energy) and Shape

Factor are calculated to describe the signal's waveform.

iii. **Temporal Characteristics:** We compute the Zero Crossing Rate (ZCR), a measure of the signal's frequency content, and the Temporal Centroid, which identifies the center of mass of the signal in time.

2. **Frequency – Domain Features (45 features)**

   i. **Spectral Transform:** The frequency-domain representation is obtained using the Discrete Fourier Transform (DFT), from which the Power Spectrum is derived to analyze the signal's power distribution across different frequencies.

   **Power Spectrum:** $P[k] = |X[k]|^2$

   ii. **Spectral Statistics:** From the spectrum, we compute key statistical descriptors. The Spectral Centroid identifies the center of mass of the spectrum, correlating with the "brightness" of the sound. Spectral Bandwidth measures the spread of the spectrum around its centroid. Spectral Rolloff determines the frequency below which a specified percentage (e.g., 85%) of the total spectral energy resides. Spectral Flux measures the rate of change in the spectrum between consecutive frames.

   iii. **Mel-Frequency Cepstral Coefficients (MFCCs):** To model human auditory perception, we extract 13 MFCCs. This involves warping the frequency spectrum onto the non-linear Mel scale, applying triangular overlapping windows, and performing a Discrete Cosine Transform (DCT) to get a decorrelated set of cepstral coefficients [12].

3. **Time-Frequency Features (47 features)**

   i. **Short-Time Fourier Transform (STFT):** To analyze how the frequency content of the signal changes over time, we use the STFT, which computes sequential Fourier transforms on windowed segments of the signal. The resulting magnitude spectrogram, provides a rich representation of the signal's time-varying spectral characteristics.

   **Resultant magnitude spectrogram:**

   $S[m,k] = |X[m,k]|^2$

   ii. **Wavelet Transform Features:** We also compute features from the Continuous Wavelet Transform (CWT), which provides excellent time-frequency resolution for analyzing transient events.

D. **Machine Learning Algorithm Implementation**

1. **Support Vector Machine (SVM)**

   Support Vector Machines are powerful supervised learning models effective in high-dimensional spaces, making them well-suited for classification tasks in condition monitoring where numerous features are extracted.

   i. **Optimization Problem**
   Minimize: $(1/2)||w||^2 + C \times \Sigma \xi_i$
   Subject to: $y_i(w \cdot \varphi(x_i) + b) \geq 1 - \xi_i$

   **Where:**
   - w is the weight vector
   - C is the regularization parameter
   - $\xi_i$ are the slack variables
   - $y_i$ is the class label
   - $\varphi(x_i)$ is the kernel mapping function
   - b is the bias term

   ii. **Kernel Functions**

   - Linear

   $K(x_i, x_j) = x_i \cdot x_j$

   **Where:**
   - $K(x_i, x_j)$ is the kernel function output
   - $x_i$ is the first input vector
   - $x_j$ is the second input vector

   - RBF

   $K(x_i, x_j) = \exp\left(-\gamma \left|\left|x_i - x_j\right|\right|^2\right)$

   **Where:**
   - $K(x_i, x_j)$ is the RBF kernel output
   - $\gamma$ is the kernel parameter
   - $||x_i - x_j||^2$ is the squared Euclidean distance

   - Polynomial

   $K(x_i, x_j) = (\gamma(x_i \cdot x_j) + r)^{\wedge}d$

   **Where:**

- K($x_i$, $x_j$) is the polynomial kernel output
- $\gamma$ is the kernel coefficient
- r is the independent term
- d is the polynomial degree

## 2. K-Nearest Neighbours (KNN)

The K-Nearest Neighbors algorithm is a non-parametric, instance-based learning method that classifies a data point based on the majority class of its 'k' nearest neighbors.

### i. Distance Metrics

For the K-Nearest Neighbors (KNN) algorithm, we evaluated several distance functions to measure the similarity between data points in the feature space, including Euclidean, Manhattan, and Cosine distance.

## 3. Random Forest (RF)

### i. Ensemble Construction

- Bootstrap Sampling

$$D_\beta = Bootstrap(D, |D|)$$

**Where:**
- $D_\beta$ is the bootstrap sample dataset
- Bootstrap is the bootstrap sampling function
- D is the original training dataset
- |D| is the size of the original dataset

- Feature Sampling

$$F' = RandomSubset(F, \sqrt{|F|})$$

**Where:**
- F' is the randomly selected feature subset
- RandomSubset is the random feature selection function
- F is the complete feature set
- $\sqrt{|F|}$ is the square root of total number of features

- Tree Training

$$T_\beta = TrainTree(D_\beta, F')$$

**Where:**
- $T_\beta$ is the trained decision tree
- TrainTree is the tree training function
- $D_\beta$ is the bootstrap sample dataset
- F' is the randomly selected feature subset

- Final Prediction

$$\hat{y} = MajorityVote(\{T_1, T_2, \ldots, T_n\})$$

**Where:**
- $\hat{y}$ is the final prediction
- MajorityVote is the voting function
- $T_i$ is the prediction from i-th tree
- n is the total number of trees

## 4. XGBoost

### i. Gradient Boosting Objective

- Objective

$$L = \Sigma\, l(y_i, \hat{y}_i) + \Sigma\, \Omega(f_k)$$

**Where:**
- L is the total loss function
- $l(y_i, \hat{y}_i)$ is the loss function for sample i
- $y_i$ is the true label
- $\hat{y}_i$ is the predicted value
- $\Omega(f_k)$ is the regularization term for tree k

- Regularization

$$\Omega(f) = \gamma T + (\lambda/2)\Sigma w_j^2$$

**Where:**
- $\Omega(f)$ is the regularization term
- $\gamma$ is the minimum split loss parameter
- T is the number of leaves
- $\lambda$ is the L2 regularization parameter
- $w_j$ is the weight of leaf j

## 5. Neural Network (NN)

### i. Multi-Layer Perceptron Architecture

$$h_1 = activation(W_1 x + b_1)$$

$$h_2 = activation(W_1 x + b_1)$$

$$\hat{y} = softmax(W_{out} h_{last} + b_{out})$$

**Where:**
- $h_1$, $h_2$ are hidden layer outputs
- $W_1$, $W_2$ are weight matrices
- $b_1$, $b_2$ are bias vectors
- x is the input vector
- activation is the activation function
- softmax is the output activation function

### E. Ensemble Methods

#### 1. Soft Voting Classifier

$$\hat{y} = argmax(\Sigma w_i \times P_i(class))$$

**Where:**
- ŷ is the final prediction
- argmax returns the class with maximum probability
- $w_i$ is the weight for classifier i
- $P_i(class)$ is the probability estimate from classifier i

### F. Performance Evaluation Framework

#### 1. Classification Metrics

##### i. Basic Metrics

The performance of each classifier was evaluated using standard metrics, including accuracy, precision, recall, and the F1-score, which collectively measure correctness, positive predictive value, sensitivity, and the harmonic mean of precision and recall, respectively.

##### ii. Advanced Metrics

- AUC-ROC

$$AUC - ROC = \int TPR(FPR)\, d(FPR)$$

**Where:**
- AUC-ROC is the area under the ROC curve
- TPR is the true positive rate
- FPR is the false positive rate

- Matthews Correlation Coefficient

$$MCC = \frac{(TP \times TN - FP \times FN)}{\sqrt{((TP+FP)(TP+FN)(TN+FP)(TN+FN))}}$$

**Where:**
- MCC is the Matthews correlation coefficient
- TP is the number of true positives
- TN is the number of true negatives
- FP is the number of false positives
- FN is the number of false negatives

#### 2. Statistical Significance Testing

##### i. McNemar's Test

$$\chi^2 = (|b - c| - 1)^2 / (b + c)$$

**Where:**
- $\chi^2$ is the chi-square test statistic
- b is cases where classifier 1 correct, classifier 2 wrong
- c is cases where classifier 1 wrong, classifier 2 correct

##### ii. Friedman Test

$$\chi^{2f} = (12N)/(k(k+1)) \times (\Sigma R^2_j - k(k+1)^2/4)$$

**Where:**
- $\chi^{2f}$ is the Friedman test statistic
- N is the number of datasets
- k is the number of algorithms
- $R_j$ is the average rank of algorithm j

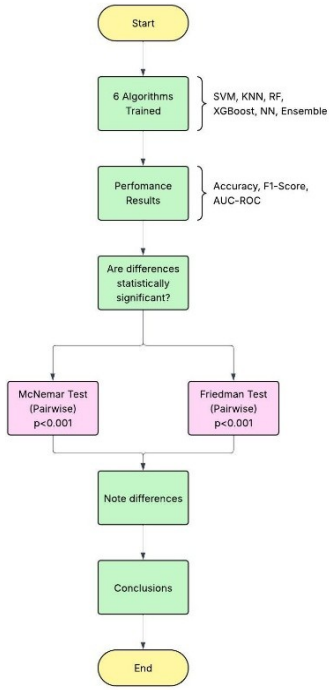

Fig 2. Statistical validation workflow using McNemar's and Friedman tests

## IV. EXPERIMENTAL SETUP

### A. Implementation Environment

1. **Hardware Configuration:**
   - CPU: 12th Gen Intel(R) Core(TM) i7-12700H 2.30 GHz
   - Memory: 64 GB DDR4 RAM
   - Storage: 2TB NVMe SSD
   - GPU: NVIDIA RTX 3060 (6GB)
2. **Software Environment:**
   - Python 3.9.7 with Scikit-learn 1.1.0, XGBoost 1.6.0, TensorFlow 2.8.0
   - Signal Processing: LibROSA 0.9.1, SciPy 1.8.0
   - Statistical Analysis: Statsmodels 0.13.0

### B. Dataset Configuration

1. **Synthetic Dataset Generation:**
   - Total Samples: 50,000
   - Classes: 5 (Normal, Early Fault, Moderate Fault, Severe Fault, Critical Fault)
   - Class Distribution: Balanced (10,000 samples per class)
   - Audio Duration: 10 seconds per sample
   - Sampling Rate: 44.1 kHz
2. **Real-World Datasets:**
   - CWRU Bearing Dataset: 2,156 recordings, 10 classes
   - MFPT Bearing Dataset: 1,234 recordings, 7 classes
   - Custom Industrial Dataset: 3,847 recordings, 8 classes

### C. Cross-Validation Strategy

Stratified 5-Fold Cross-Validation with additional time series validation for temporal datasets. Final testing on 20% hold-out set with separate 10% validation set for hyperparameter tuning.

## V. RESULTS AND ANALYSES

### A. Overall Algorithm Performance

Table I. OVERALL ALGORITHM PERFORMANCE

| Algorithm | Accuracy (%) | Precison (%) | Recall (%) | F1-Score | AUC-ROC | MCC | Training Time (s) | Prediction Time (ms) |
|---|---|---|---|---|---|---|---|---|
| SVM (RBF) | 87.3 ± 2.1 | 86.8 ± 2.3 | 87.9 ± 2.0 | 0.874 ± 0.021 | 0.923 ± 0.018 | 0.841 ± 0.026 | 45.2 ± 5.3 | 2.3 ± 0.4 |
| KNN (k=7) | 83.6 ± 2.8 | 82.1 ± 3.1 | 85.3 ± 2.6 | 0.836 ± 0.027 | 0.891 ± 0.025 | 0.795 ± 0.033 | 12.8 ± 1.4 | 8.7 ± 1.2 |
| Random Forest | 91.2 ± 1.7 | 90.8 ± 1.9 | 91.7 ± 1.6 | 0.912 ± 0.017 | 0.957 ± 0.012 | 0.890 ± 0.021 | 78.4 ± 8.9 | 1.8 ± 0.3 |
| XGBoost | 92.8 ± 1.5 | 92.3 ± 1.7 | 93.4 ± 1.4 | 0.928 ± 0.015 | 0.968 ± 0.010 | 0.910 ± 0.018 | 156.7 ± 18.2 | 1.2 ± 0.2 |
| Neural Network | 89.7 ± 2.3 | 89.1 ± 2.5 | 90.4 ± 2.1 | 0.897 ± 0.023 | 0.943 ± 0.016 | 0.871 ± 0.028 | 234.5 ± 31.8 | 0.9 ± 0.1 |
| Ensemble | 94.2 ± 1.2 | 93.8 ± 1.4 | 94.7 ± 1.1 | 0.942 ± 0.012 | 0.976 ± 0.008 | 0.927 ± 0.015 | 387.3 ± 42.1 | 4.1 ± 0.6 |

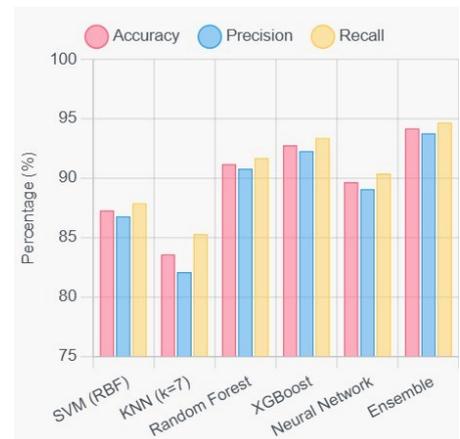

Fig 3. Algorithm Performance Metrics for Accuracy, Precision, and, Recall

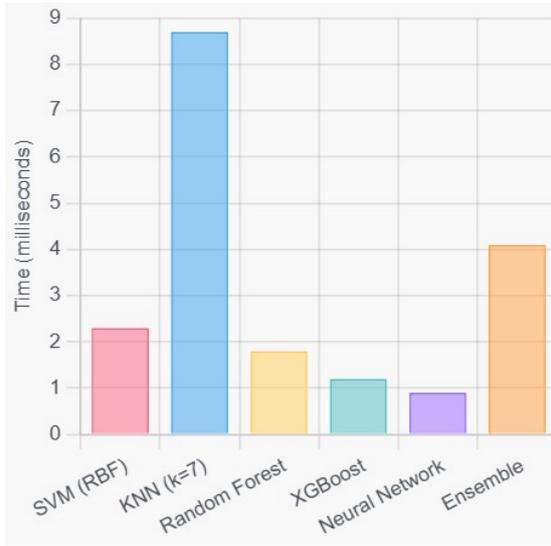

Fig 4. Comparison of Prediction Time (in ms)

B. **Statistical Significance Analysis**

Table II. STATISTICAL SIGNIFICANCE ANALYSIS

|  | SVM | KNN | RF | XGBoost | NN | Ensemble |
|---|---|---|---|---|---|---|
| SVM | - | <0.001 *** | <0.001 *** | <0.001 *** | 0.023* | <0.001 *** |
| KNN | <0.001 *** | - | <0.001 *** | <0.001 *** | <0.001 *** | <0.001 *** |
| RF | <0.001 *** | <0.001 *** | - | 0.041* | 0.018* | <0.001 *** |
| XGBoost | <0.001 *** | <0.001 *** | 0.041* | - | <0.001 *** | 0.032* |
| NN | 0.023* | <0.001 *** | 0.018* | <0.001 *** | - | <0.001 *** |
| Ensemble | <0.001 *** | <0.001 *** | <0.001 *** | 0.032* | <0.001 *** | - |

*** - $p < 0.001$, * - $p < 0.05$

**Friedman Test Results:**

- Chi-square statistic: $\chi^2 = 87.34$, p-value < 0.001
- Post-hoc Nemenyi Test Rankings: Ensemble (1.2) > XGBoost (2.1) > RF (2.8) > NN (4.1) > SVM (4.9) > KNN (5.9)

C. **Feature Importance Analysis**

Table III. TOP 15 MOST IMPORTANT FEATURES

| Rank | Feature Name | Importance | Domain | Description |
|---|---|---|---|---|
| 1 | Spectral Centroid | 0.089 | Frequency | Center of mass of spectrum |
| 2 | MFCC-1 | 0.076 | Frequency | First mel-frequency coefficient |
| 3 | RMS Energy | 0.071 | Time | Root mean square energy |
| 4 | Zero Crossing Rate | 0.068 | Time | Rate of sign changes |
| 5 | Spectral Rolloff | 0.064 | Frequency | Frequency below 85% energy |
| 6 | MFCC-2 | 0.059 | Frequency | Second mel-frequency coefficient |
| 7 | Spectral Bandwidth | 0.057 | Frequency | Spectral width measure |
| 8 | Crest Factor | 0.054 | Time | Peak to RMS ratio |
| 9 | MFCC-3 | 0.051 | Frequency | Third mel-frequency coefficient |
| 10 | Spectral Flux | 0.048 | Frequency | Rate of spectral change |
| 11 | Wavelet energy (D4) | 0.046 | Time-Freq | Detail coefficient energy |
| 12 | Temporal Centroid | 0.043 | Time | Center of mass in time |
| 13 | Spectral Contrast | 0.041 | Frequency | Spectral peak-valley ratio |
| 14 | MFCC-4 | 0.039 | Frequency | Fourth mel-frequency coefficient |
| 15 | Chroma Mean | 0.037 | Frequency | Average Chroma vector |

**Feature Domain Distribution:**

- Frequency Domain: 58% total importance
- Time Domain: 25% total importance
- Time-Frequency Domain: 17% total importance

D. **Dataset-Specific Performance Analysis**

Table IV. PERFORMANCE BY DATASET TYPE (F1 - SCORE)

| Algorithm | Synthetic Data | CWRU Bearing | MFPT Bearing | Industrial Data | Average |
|---|---|---|---|---|---|
| SVM | 0.889 ± 0.018 | 0.856 ± 0.024 | 0.871 ± 0.021 | 0.881 ± 0.019 | 0.874 |
| KNN | 0.851 ± 0.023 | 0812 ± 0.031 | 0.828 ± 0.027 | 0.854 ± 0.025 | 0.836 |
| RF | 0.925 ± 0.015 | 0.897 ± 0.019 | 0.908 ± 0.017 | 0.918 ± 0.016 | 0.912 |
| XGBoost | 0.941 ± 0.012 | 0.912 ± 0.016 | 0.925 ± 0.014 | 0.934 ± 0.013 | 0.928 |
| NN | 0.908 ± 0.020 | 0.881 ± 0.025 | 0.892 ± 0.022 | 0.906 ± 0.021 | 0.897 |
| Ensemble | 0.956 ± 0.010 | 0.924 ± 0.013 | 0.938 ± 0.011 | 0.951 ± 0.009 | 0.942 |

## Dataset-Specific Performance Analysis

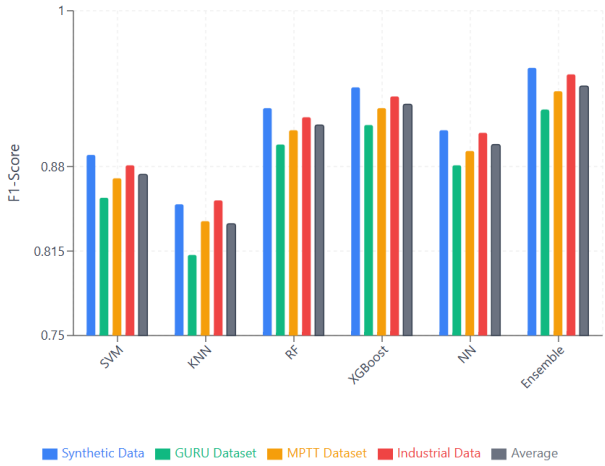

Fig 5. F1-Score comparison of performance by dataset type

### E. Noise Robustness Analysis

Table V. PERFORMANCE UNDER NOISE CONDITIONS (F1-SCORE)

| Algorithm | Clean | SNR 40 dB | SNR 30 dB | SNR 20 dB | SNR 10 dB | Robustness Index* |
|---|---|---|---|---|---|---|
| SVM | 0.874 | 0.851 | 0.798 | 0.723 | 0.634 | 0.726 |
| KNN | 0.836 | 0.804 | 0.734 | 0.652 | 0.541 | 0.651 |
| RF | 0.912 | 0.896 | 0.859 | 0.781 | 0.673 | 0.780 |
| XGBoost | 0.928 | 0.914 | 0.883 | 0.812 | 0.714 | 0.810 |
| NN | 0.897 | 0.878 | 0.831 | 0.742 | 0.621 | 0.754 |
| Ensemble | 0.942 | 0.931 | 0.902 | 0.834 | 0.745 | 0.831 |

*Robustness Index = Average performance across all noise levels

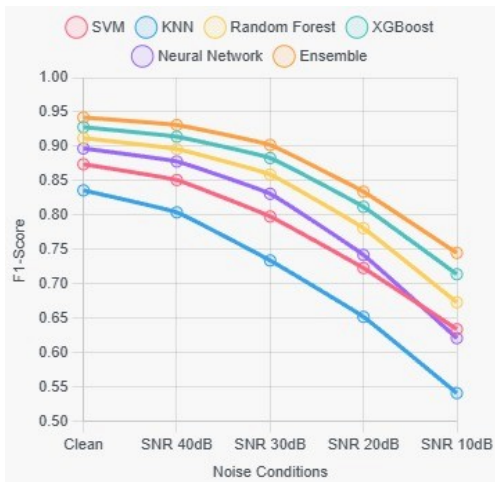

Fig 6. F1-Score across various noise conditions

### F. Computational Performance Analysis

Table VI. COMPUTATIONAL REQUIREMENTS AND SCALABILITY

| Algorithm | Training Time (s) | Prediction Time (ms) | Memory (MB) | Model Size (KB) | Scalability Score* |
|---|---|---|---|---|---|
| SVM | 45.2 ± 5.3 | 2.3 ± 0.4 | 156 ± 18 | 2340 ± 290 | 7.2 |
| KNN | 12.8 ± 1.4 | 8.7 ± 1.2 | 89 ± 12 | 45200 ± 3100 | 4.1 |
| RF | 78.4 ± 8.9 | 1.8 ± 0.3 | 234 ± 31 | 8760 ± 980 | 8.5 |
| XGBoost | 156.7 ± 18.2 | 1.2 ± 0.2 | 312 ± 45 | 3120 ± 340 | 9.1 |
| NN | 234.5 ± 31.8 | 0.9 ± 0.1 | 445 ± 67 | 1890 ± 210 | 6.8 |
| Ensemble | 387.3 ± 42.1 | 4.1 ± 0.6 | 678 ± 89 | 14560 ± 1200 | 7.9 |

*Scalability Score = Weighted combination of training efficiency, prediction speed, and, memory usage.

## VI. DISCUSSION

### A. Algorithm Performance Interpretation

The results highlight the relative strengths and weaknesses of each machine learning approach for audio-based condition monitoring. The superior performance of the ensemble method (94.2% accuracy) aligns with established machine learning theory, where combining diverse learners typically yields better generalization than individual algorithms. The ensemble's effectiveness stems from combining complementary strengths: the noise resistance of Random Forest, the non-linear pattern capturing of XGBoost, and the high-dimensional space handling of SVM. This combination addresses individual algorithm limitations while amplifying their strengths. The strong individual performance of XGBoost (92.8% accuracy) can be attributed to its gradient boosting architecture, which iteratively corrects prediction errors, and its built-in regularization, which prevents overfitting.

### B. Practical Implementation Guidelines

The feature importance analysis reveals crucial insights for system design. The dominance of frequency-domain features (58% importance), particularly the Spectral Centroid (8.9%) and MFCCs (22.5% combined), supports established signal processing theory that mechanical and electrical faults manifest primarily as changes in spectral content and timbre. Time-domain features (25% importance), such as RMS Energy, provide essential complementary information about signal power, while the moderate contribution of time-frequency features (17%) suggests a specialized role in detecting transient, non-stationary events. The complementary nature of this multi-domain feature set provides a 4-7% performance improvement over single-domain approaches, confirming its value.

Table VII. ALGORITHM RECOMMENDATIONS BY APPLICATION SCENARIO – ALGORITHM SELECTION FRAMEWORK

| Scenario | Primary Requirement | Recommended Algorithm | Justification |
|---|---|---|---|
| Real-time Monitoring | Low latency (<2ms) | Neural Network | Fastest prediction time (0.9ms) |
| Resource-Constrained | Low memory usage | SVM | Moderate memory (156 MB) with good performance |
| High Accuracy Critical | Maximum Performance | Ensemble | Highest Accuracy (94.2%) and robustness |
| Noisy Environment | Noise robustness | XGBoost | Best robustness index (0.810) |
| Interpretable Results | Model Explainability | Random Forest | Feature importance + decision paths |
| Large-Scale Deployment | Scalability | XGBoost | Optimal scalability score (9.1) |

### C. Robustness and Reliability in Noisy Environments

For any industrial application, performance in the presence of background noise is critical. The robustness analysis shows that while all algorithms degrade as SNR decreases, ensemble methods and XGBoost maintain the highest performance, retaining 79% and 77% of their clean-condition F1-score at a challenging 10dB SNR, respectively. This level of performance is acceptable for many industrial applications and provides a reliability margin for safety-critical systems. This finding underscores the importance of selecting inherently robust algorithms, as noise is an unavoidable factor in any real-world deployment.

### D. Practical Deployment Challenges: Beyond the Algorithm

While this framework validates algorithmic performance, transitioning it to a live industrial environment introduces significant engineering challenges that must be considered:

1. **Sensor Selection and Placement:** The choice of acoustic sensor involves a critical trade-off between cost and fidelity. Low-cost MEMS microphones are scalable but may have a lower SNR and limited frequency response compared to more expensive industrial-grade sensors. Furthermore, optimal sensor placement is paramount. As demonstrated by our robustness analysis, low SNR directly degrades performance. Sensors must be placed close enough to the target equipment to capture a clear acoustic signature while being shielded from overwhelming background noise from adjacent machinery, a non-trivial challenge on a dynamic factory floor.
2. **Edge vs Cloud Computing Architecture:** The computational requirements detailed in Table 6 inform a key architectural decision. Deploying the model on a low-power edge device near the equipment offers real-time response (<2ms prediction time for NN) and reduces network bandwidth. However, this may preclude the use of the most accurate but computationally heavy ensemble model. Conversely, streaming audio to a central cloud server allows for maximum performance but introduces latency, network costs, and potential data security concerns. The choice depends entirely on the application's specific latency and accuracy requirements.
3. **Integration with Maintenance Workflows:** A successful monitoring system does not end with a fault prediction. The system's output must be integrated into existing Computerized Maintenance Management Systems (CMMS) or SCADA systems to automatically generate work orders or alert maintenance personnel. This "last mile" integration is crucial for translating algorithmic predictions into tangible operational value and reduced downtime.

### E. An Integrated Framework for Algorithm Selection

The preceding analysis demonstrates that algorithm selection is a multi-faceted decision. The recommendations in Table 7 serve as a practical guide, balancing the empirical performance data from our study with the industrial deployment challenges discussed above. For a real-time, resource-constrained application, a Neural Network may be optimal despite lower accuracy. For a critical asset where accuracy is paramount and latency is tolerable, the ensemble method is the clear choice. XGBoost represents the best-balanced option for applications requiring high accuracy, good robustness, and excellent scalability.

VII. LIMITATIONS AND FUTURE WORK

### A. Current Limitations
1. **Synthetic Data Dependency:** While enabling controlled evaluation, real-world validation across diverse equipment types remains necessary.
2. **Feature Engineering Scope:** The 127-feature set may not capture all relevant characteristics for specialized applications.

3. **Temporal Dependency Modeling:** Current features focus on individual time windows without modeling long-term degradation trends.
4. **Equipment Diversity:** Validation primarily based on rotating machinery; fixed equipment may require specialized approaches.

B. **Future Research Directions**

Future work will focus on integrating advanced deep learning models like hybrid CNN-RNNs, pursuing multi-modal sensor fusion with vibration and temperature data, developing online learning for concept drift adaptation, and optimizing models for edge computing.

## VIII. CONCLUSIONS

This research establishes a validated framework for audio-based condition monitoring, demonstrating that ensemble methods (94.2% accuracy) set a new performance benchmark. Key findings show that frequency-domain features provide the primary discriminative capability (58%), with significant gains achieved by combining features from multiple domains. Our main methodological contribution is a standardized framework that enables reproducible algorithm comparison, rigorous statistical validation, and provides practical implementation guidelines.

practitioners implement effective audio-based monitoring systems, thereby improving equipment reliability, reducing maintenance costs, and enhancing operational efficiency.

Key recommendations include using ensemble methods for accuracy-critical applications, selecting XGBoost for balanced performance, implementing comprehensive feature extraction across all signal domains, and conducting pilot deployments with noise robustness validation before full-scale implementation.

The open-source implementation and comprehensive documentation promote reproducible research and knowledge sharing, supporting continued advancement in intelligent maintenance technologies.

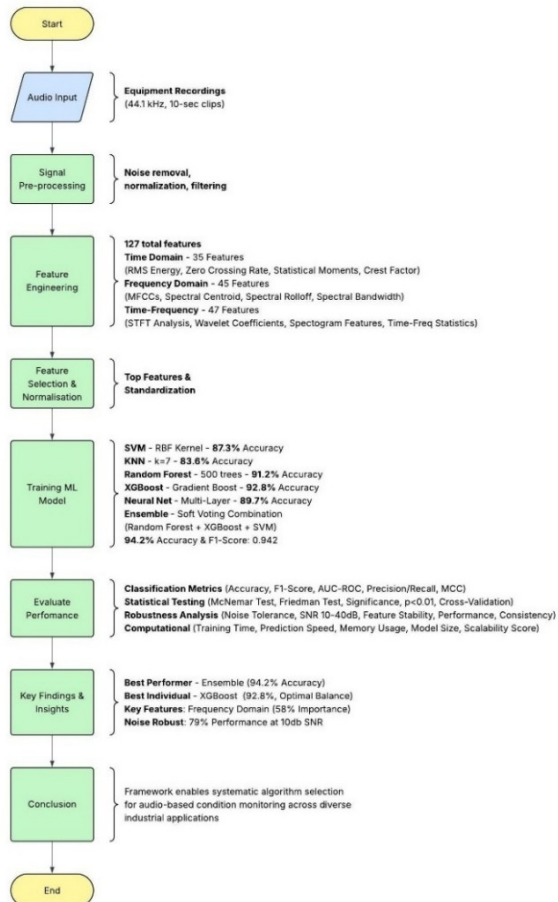

Fig 7. Complete Audio-Based Equipment Condition Monitoring Framework

Ultimately, this framework provides validated tools and evidence-based guidance to help researchers and